\documentclass{article}

\usepackage{arxiv}

\usepackage[utf8]{inputenc}
\usepackage[T1]{fontenc}
\usepackage{hyperref}
\usepackage{url}
\usepackage{booktabs}
\usepackage{amsmath}
\usepackage{amsfonts}
\usepackage{nicefrac}
\usepackage{microtype}
\usepackage{graphicx}
\usepackage[numbers]{natbib}
\usepackage{doi}

\title{A Comparative Study of PyCaret AutoML and CNN-BiLSTM for Binary Hate Speech Detection in Indonesian Twitter}
\date{}
\author{Tanty Widiyastuti \\
        Department of Data Science\\
        Institut Teknologi Sumatera (ITERA)\\
        Lampung, Indonesia \\
        \texttt{tanty.123450094@student.itera.ac.id} \\
        \And%
        Mayada \\
        Department of Data Science\\
        Institut Teknologi Sumatera (ITERA)\\
        Lampung, Indonesia \\
        \texttt{mayada.121450145@student.itera.ac.id} \\
        \AND%
        Adisty Syawalda Ariyanto \\
        Department of Data Science\\
        Institut Teknologi Sumatera (ITERA)\\
        Lampung, Indonesia \\
        \texttt{adisty.121450136@student.itera.ac.id} \\
        \And%
        Luluk Muthoharoh, M.Si. \\
        Department of Data Science\\
        Institut Teknologi Sumatera (ITERA)\\
        Lampung, Indonesia \\
        \texttt{luluk.muthoharoh@sd.itera.ac.id} \\
        \AND%
        Ardika Satria, M.Si. \\
        Department of Data Science\\
        Institut Teknologi Sumatera (ITERA)\\
        Lampung, Indonesia \\
        \texttt{ardika.satria@sd.itera.ac.id} \\
        \And%
        Martin Clinton Tosima Manullang, Ph.D. \\
        Department of Informatics\\
        Institut Teknologi Sumatera (ITERA)\\
        Lampung, Indonesia \\
        \texttt{martin.manullang@if.itera.ac.id} \\
}
\renewcommand{\shorttitle}{Binary Hate Speech Detection in Indonesian Twitter}
\renewcommand{\headeright}{}
\renewcommand{\undertitle}{}

\hypersetup{
pdftitle={A Comparative Study of PyCaret AutoML and CNN-BiLSTM for Binary Hate Speech Detection in Indonesian Twitter},
pdfsubject={A comparative study of PyCaret AutoML and CNN-BiLSTM for binary hate speech detection in Indonesian Twitter},
pdfauthor={Tanty Widiyastuti, Mayada, Adisty Syawalda Ariyanto, Luluk Muthoharoh, Ardika Satria, Martin Clinton Tosima Manullang},
pdfkeywords={natural language processing, hate speech detection, Indonesian Twitter, binary classification, PyCaret AutoML, TF-IDF, CNN-BiLSTM, text classification},
}

\begin{document}
\maketitle

\begin{abstract}
This paper compares a PyCaret AutoML branch and a CNN-BiLSTM branch for binary hate speech detection on Indonesian Twitter using the HS label from the corpus of \citet{ibrohim2019}. Both branches share the same preprocessing pipeline so that the comparison reflects modelling differences rather than inconsistent data preparation. The conventional branch uses TF-IDF with a lexicon-based abusive-word count, whereas the neural branch learns dense token representations and captures both local phrase patterns and bidirectional context. The benchmark is built from the released 13{,}130-row annotation table, whose HS label yields a 58:42 class ratio. On the held-out split, CNN-BiLSTM achieves the best result with 83.8\% accuracy, 79.8\% precision, 82.7\% recall, and 81.2\% F1-score. Within the PyCaret branch, Random Forest is the strongest conventional model with 77.2\% accuracy and 77.0\% F1-score. The neural branch therefore improves accuracy by 6.6 points and F1-score by 4.2 points. Exploratory corpus analysis, learning curves, and confusion matrices show that the dataset is short-text, moderately imbalanced, and still difficult because many decisions depend on local lexical cues plus short contextual composition. The study concludes that PyCaret AutoML is an effective conventional benchmarking framework, whereas CNN-BiLSTM is the stronger end model for the reported benchmark setting.
\end{abstract}


\section{Introduction}
Automatic hate speech detection on Indonesian Twitter is difficult because tweets are short, noisy, informal, and strongly context-dependent. Harmful intent is not always expressed by a single slur; it may emerge from target reference, phrase composition, negation, or pragmatic framing. This makes the task a useful setting for comparing feature-based machine learning and sequence-aware deep learning.

This paper studies the problem as a controlled benchmark. Rather than comparing CNN-BiLSTM against one arbitrarily chosen classical classifier, the conventional branch is organized through PyCaret AutoML. PyCaret is used here as a benchmarking layer that standardizes model comparison over a shared feature space and helps surface strong conventional candidates under the same experimental design. The deep-learning branch uses CNN-BiLSTM because Indonesian Twitter hate speech often depends on both short local fragments and nearby sequential context.

The benchmark focuses on the hate-speech (HS) label from the released corpus of \citet{ibrohim2019}. Both branches begin from the same cleaned corpus and are evaluated on the same held-out split with the same core metrics. This shared design makes the comparison easier to interpret because performance differences are less likely to be caused by different preprocessing assumptions.

The contribution of the paper is fourfold. First, it reformulates the released multi-label corpus into a controlled binary HS benchmark. Second, it operationalizes the conventional branch through PyCaret AutoML instead of through a single weak baseline. Third, it evaluates a compact CNN-BiLSTM architecture tailored to short contextual text. Fourth, it interprets the benchmark using corpus statistics, final metrics, training curves, and confusion matrices. The central question is therefore precise: after the conventional branch is strengthened through AutoML, does CNN-BiLSTM still provide a meaningful advantage?

That question can be restated as three practical research questions. Can a PyCaret-based conventional branch built on strong lexical features remain competitive on Indonesian hate speech? Does CNN-BiLSTM still outperform the strongest PyCaret candidate when preprocessing and evaluation are held constant? If the neural branch does outperform the conventional branch, is the gain visible only in headline accuracy, or also in predictive balance, learning dynamics, and residual error patterns? Framing the study this way helps connect the benchmark to realistic moderation-oriented use cases rather than to a narrow model-versus-model comparison.

\section{Related Work}
The Indonesian hate-speech corpus introduced by \citet{ibrohim2019} remains the main benchmark foundation for this task. Its annotation scheme distinguishes hate speech from abusive language, which is important because explicit abuse is informative but not identical to hate speech. This distinction motivates the present comparison between lexical and contextual modelling.

Lexicon-aware conventional models remain relevant in Indonesian toxic-language detection. Abusive-word information can improve Indonesian hate-speech classification~\citep{pamungkas2024}, while TF-IDF remains a strong sparse representation for short text~\citep{ramos2003}. In this paper, PyCaret is useful because it standardizes comparison across multiple conventional estimators under the same workflow~\citep{hutter2019,pycaretdocs}.

Neural text classification addresses limitations of purely sparse lexical features. CNNs are effective for learning local n-gram patterns~\citep{kim2014}, and bidirectional recurrent models help encode left and right context~\citep{hochreiter1997,schuster1997}. For Indonesian Twitter, this is particularly relevant because short utterances often depend on both local abusive fragments and nearby contextual framing. Transformer-based Indonesian models such as IndoBERT and IndoBERTweet further show the value of contextual representation~\citep{koto2020,koto2021}, but the present paper asks a narrower question: whether a compact CNN-BiLSTM still outperforms a strong AutoML-managed conventional branch under shared benchmark conditions.

This focus also matters from a benchmarking perspective. If the conventional branch were weak, any neural gain would be difficult to interpret. PyCaret is therefore valuable because it reduces implementation inconsistency, standardizes the comparison workflow, and makes the classical side of the benchmark more credible before the deeper architecture is evaluated. The value of the present comparison lies precisely in clarifying what is gained when moving from a strong AutoML-managed lexical branch to a sequence-aware neural branch on the same Indonesian Twitter task.

\section{Dataset and Preprocessing}
\subsection{Corpus Characteristics}
The experiments use the Indonesian Twitter hate-speech corpus of \citet{ibrohim2019}. The released source corpus is reported as containing 13{,}169 tweets, while the released annotation table used here contains 13{,}130 rows. All counts, splits, and metrics in this paper are based on that 13{,}130-row table.

The benchmark is formed by taking HS as the target label, with HS$=1$ treated as hate speech and HS$=0$ treated as non-hate speech. In this benchmark, the HS label contains 7{,}577 non-HS tweets and 5{,}553 HS tweets, or about 58:42. The abusive-language label contains 8{,}095 non-abusive tweets and 5{,}035 abusive tweets, or about 62:38. Tweets are short and right-skewed in length, with an average of about 14.5 words and a large majority below 20 words. These properties motivate the use of precision, recall, and F1-score in addition to accuracy and also support a model that can capture short contextual patterns.

The released package is important for more than the annotation table alone. It also provides normalization resources and an abusive lexicon that make later Indonesian hate-speech research easier to reproduce. Usernames and URLs are already anonymized as \texttt{USER} and \texttt{URL}, which supports a consistent preprocessing workflow across studies. The present benchmark therefore treats the released corpus as both a labelled dataset and a reusable resource package for normalization and lexicon-aware modelling.

\begin{figure}[t]
    \centering
    \includegraphics[width=0.90\linewidth]{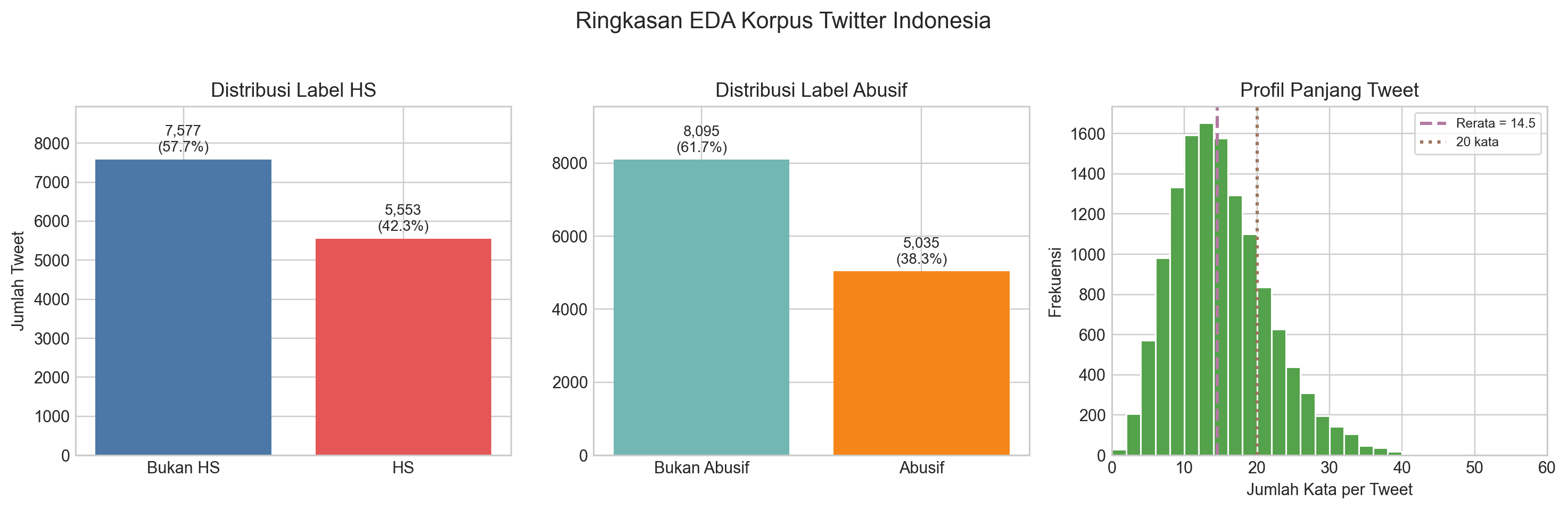}
    \caption{EDA overview of the benchmark corpus: HS distribution, abusive-language distribution, and tweet-length profile.}\label{fig:eda}
\end{figure}

Figure~\ref{fig:eda} summarizes the benchmark through HS distribution, abusive-language distribution, and tweet-length statistics. The plot labels are shown in Indonesian, while the text of the paper uses English class descriptions for consistency.

\subsection{Preprocessing Strategy}
The preprocessing pipeline is shared across both branches. Tweets are lowercased, URLs and user mentions are removed, and extraneous symbols are normalized. Slang normalization uses the mappings distributed with the corpus package, which helps reduce lexical sparsity caused by Indonesian social-media spelling variation. In practical terms, this stage reuses the released typo-and-slang mappings rather than a newly built dictionary. Stopword removal is used as a noise-reduction step, and the abusive lexicon released with the corpus is reused to compute a simple abusive-word count.

This shared design has an important benchmark purpose. The PyCaret AutoML branch and the CNN-BiLSTM branch should differ mainly in modelling capacity, not in how the text is cleaned. For that reason, the same cleaned text is used to derive both the sparse lexical representation for conventional models and the padded token sequences for the neural model. The preprocessing stage is therefore part of the comparative design itself: it gives the conventional branch access to a strong lexical baseline while preserving a fair starting point for the neural branch.

\section{Methodology}
\subsection{Problem Formulation}
The task is binary classification. Let each tweet be denoted by $x_i$ with label $y_i \in \{0,1\}$, where $y_i=1$ denotes hate speech. The neural branch predicts a probability $\hat{y}_i \in [0,1]$ and is trained with binary cross-entropy:
\begin{equation}
\mathcal{L}_{\mathrm{BCE}} = -\frac{1}{N}\sum_{i=1}^{N}\left[y_i\log \hat{y}_i + (1-y_i)\log(1-\hat{y}_i)\right].
\end{equation}
This formulation keeps the comparison focused on one moderation-relevant question: should a tweet be classified as hate speech or not?

\subsection{PyCaret AutoML Pipeline}
In this paper, PyCaret is not treated as a classifier in its own right; it is the workflow that standardizes the conventional branch. After shared cleaning, each tweet is represented by TF-IDF plus an abusive-word count. The TF-IDF weight for term $t$ in document $d$ is written as
\begin{equation}
\mathrm{TFIDF}(t,d) = \mathrm{tf}(t,d) \cdot \log \frac{N}{\mathrm{df}(t)+1},
\end{equation}
and the abusive feature is defined as
\begin{equation}
f_{\mathrm{abusive}}(x_i) = \sum_{w \in x_i} \mathbf{1}(w \in \mathcal{V}_{\mathrm{abusive}}),
\end{equation}
so that the final feature vector is
\begin{equation}
\mathbf{z}_i = [\mathbf{v}_i ; f_{\mathrm{abusive}}(x_i)],
\end{equation}
where $\mathbf{v}_i$ is the TF-IDF vector and $\mathcal{V}_{\mathrm{abusive}}$ denotes the abusive lexicon.

The PyCaret workflow compares multiple estimators over this shared representation and ranks them under a consistent scoring setup~\citep{pycaretdocs}. In generic form, if $\mathcal{M}$ denotes the set of candidate conventional models and $\mathrm{F1}^{(k)}(m)$ denotes the F1-score of model $m$ on fold $k$, the ranking target can be written as
\begin{equation}
m^{\ast} = \arg\max_{m \in \mathcal{M}} \frac{1}{K}\sum_{k=1}^{K} \mathrm{F1}^{(k)}(m).
\end{equation}
In practice, the reported representation uses a maximum TF-IDF vocabulary of 5{,}000 with unigram and bigram features plus the abusive-count feature.

For substantive reporting, the paper retains three representative conventional models: linear SVM, Naive Bayes, and Random Forest. These models cover linear, probabilistic, and nonlinear ensemble perspectives, making the conventional branch stronger than a single ad hoc baseline. SVM tests how far a margin-based linear separator can go in sparse space, Naive Bayes provides a lightweight probabilistic reference, and Random Forest captures nonlinear interaction among lexical cues. In the reported benchmark, Random Forest emerges as the strongest PyCaret model.

\subsection{Deep Learning Architecture}\label{subsec:deep-learning}
The deep-learning branch uses a CNN-BiLSTM architecture because the task depends on both local lexical fragments and short contextual interaction. After tokenization and padding to a maximum length of 50, each word is mapped to a 100-dimensional trainable embedding. A convolutional block extracts short phrase features,
\begin{equation}
\mathbf{c}_t = \phi\left(\mathbf{W}_c[\mathbf{e}_t;\dots;\mathbf{e}_{t+k-1}] + \mathbf{b}_c\right),
\end{equation}
with kernel size $k=3$. These features are then encoded by a bidirectional LSTM to form the contextual state
\begin{equation}
\mathbf{h}_t = [\overrightarrow{\mathbf{h}}_t ; \overleftarrow{\mathbf{h}}_t],
\end{equation}
and the final tweet representation is mapped to the output probability
\begin{equation}
\hat{y}_i = \sigma(\mathbf{w}^{\top}\mathbf{h}^{\ast}_i + b).
\end{equation}
At the gate level, the recurrent unit updates input, forget, and output controls so that useful context can be retained while noisy signals are suppressed:
\begin{align}
\mathbf{i}_t &= \sigma(\mathbf{W}_i\mathbf{c}_t + \mathbf{U}_i\mathbf{h}_{t-1} + \mathbf{b}_i), \\
\mathbf{f}_t &= \sigma(\mathbf{W}_f\mathbf{c}_t + \mathbf{U}_f\mathbf{h}_{t-1} + \mathbf{b}_f), \\
\mathbf{o}_t &= \sigma(\mathbf{W}_o\mathbf{c}_t + \mathbf{U}_o\mathbf{h}_{t-1} + \mathbf{b}_o).
\end{align}

The architectural motivation is straightforward. CNN captures short fragments such as abusive phrases, target-bearing n-grams, and compact intensifiers, while BiLSTM interprets those cues using left and right context. This is appropriate for short Indonesian tweets, where local lexical signals are important but often change force because of nearby context. The dense embedding layer is also useful because it reduces the dependence on exact surface-form overlap and lets the classifier learn task-specific token representations from the benchmark itself.

The principal neural hyperparameters are summarized in Table~\ref{tab:hyperparams}. The model is intentionally kept at a moderate scale so that gains are more plausibly linked to representational suitability than to brute-force capacity alone. With a corpus of roughly thirteen thousand tweets, a much larger network would make improvements harder to interpret because it would increase the risk of memorization.

\begin{table}[t]
    \centering
    \small
    \caption{Main CNN-BiLSTM hyperparameters.}\label{tab:hyperparams}
    \begin{tabular}{ll}
        \toprule
        Hyperparameter & Value \\
        \midrule
        Embedding dimension & 100 \\
        Sequence length & 50 \\
        CNN filters & 64 \\
        Kernel size & 3 \\
        BiLSTM units & 50 \\
        Dropout rate & 0.2 \\
        Learning rate & $10^{-3}$ \\
        Batch size & 64 \\
        \bottomrule
    \end{tabular}
\end{table}

\section{Experiments}
\subsection{Experimental Configuration}
The benchmark uses a single held-out 20\% test split, corresponding to 2{,}626 tweets, while the neural branch uses a validation subset drawn from the training portion for monitoring. The PyCaret branch compares multiple conventional estimators over the shared TF-IDF plus abusive-count representation and retains SVM, Naive Bayes, and Random Forest for reporting. The CNN-BiLSTM branch uses the hyperparameters in Table~\ref{tab:hyperparams}. For contextual comparison only, the training-curve and confusion-matrix figures also include an auxiliary abusive-label run generated from the same released benchmark table under the same architecture and matched split policy.

Because the paper reports one controlled split rather than repeated-run averages, the results should be read as benchmark-specific evidence rather than as a full variance study. Even so, the comparison remains informative because the classical branch is not underdeveloped, the preprocessing logic is aligned across both tracks, and the diagnostic figures make it possible to interpret not only the final scores but also the learning behaviour and residual error structure of the neural model.

\subsection{Evaluation Metrics}
Model quality is evaluated through accuracy, precision, recall, and F1-score. Let $TP$, $TN$, $FP$, and $FN$ denote true positives, true negatives, false positives, and false negatives. The metrics are defined as
\begin{equation}
\mathrm{Accuracy}=\frac{TP+TN}{TP+TN+FP+FN},\quad
\mathrm{Precision}=\frac{TP}{TP+FP},\quad
\mathrm{Recall}=\frac{TP}{TP+FN},
\end{equation}
\begin{equation}
\mathrm{F1}=2\cdot \frac{\mathrm{Precision}\cdot\mathrm{Recall}}{\mathrm{Precision}+\mathrm{Recall}}.
\end{equation}
Accuracy gives a global summary, but the HS benchmark is moderately imbalanced and operationally sensitive to both missed hateful tweets and over-flagged non-hateful tweets. For that reason, F1-score is treated as a key comparative measure.

\section{Results and Discussion}
Table~\ref{tab:results} summarizes the main HS benchmark. The CNN-BiLSTM achieves the best result across all reported models. Relative to PyCaret-RF, the strongest conventional comparator surfaced by the AutoML branch, the neural model improves accuracy by 6.6 points, precision by 4.0 points, recall by 4.4 points, and F1-score by 4.2 points. This gap is methodologically meaningful because the comparator is not a weak baseline chosen for convenience; it is the strongest model among the reported PyCaret candidates under the same shared representation.

Within the PyCaret branch, SVM is the weakest model, suggesting that a linear separator over sparse lexical features is not sufficient for the full HS boundary. Naive Bayes improves recall but remains weaker in precision. Random Forest is the strongest AutoML candidate, which indicates that nonlinear interaction among lexical signals matters. Even so, the CNN-BiLSTM still performs best, which supports the claim that sequence-aware modelling captures information that is not recovered fully by engineered lexical features alone.

\begin{table}[t]
    \centering
    \small
    \caption{Benchmark comparison on the HS task.}\label{tab:results}
    \begin{tabular}{lcccc}
        \toprule
        Method & Acc. & Prec. & Rec. & F1 \\
        \midrule
        PyCaret-SVM & 72.3 & 70.1 & 74.5 & 72.2 \\
        PyCaret-NB & 75.8 & 73.2 & 78.1 & 75.6 \\
        PyCaret-RF & 77.2 & 75.8 & 78.3 & 77.0 \\
        CNN-BiLSTM & 83.8 & 79.8 & 82.7 & 81.2 \\
        \bottomrule
    \end{tabular}
\end{table}

The ranking is also consistent with the linguistic structure of the corpus. Explicit abusive vocabulary remains useful, but hate speech cannot be reduced to lexical toxicity alone. The CNN layer can detect short phrase-level patterns, while the BiLSTM layer disambiguates those patterns using nearby context. This is especially helpful when abusive tokens change function because of negation, quotation, target specification, or short contextual framing.

\begin{figure}[t]
    \centering
    \includegraphics[width=0.74\linewidth]{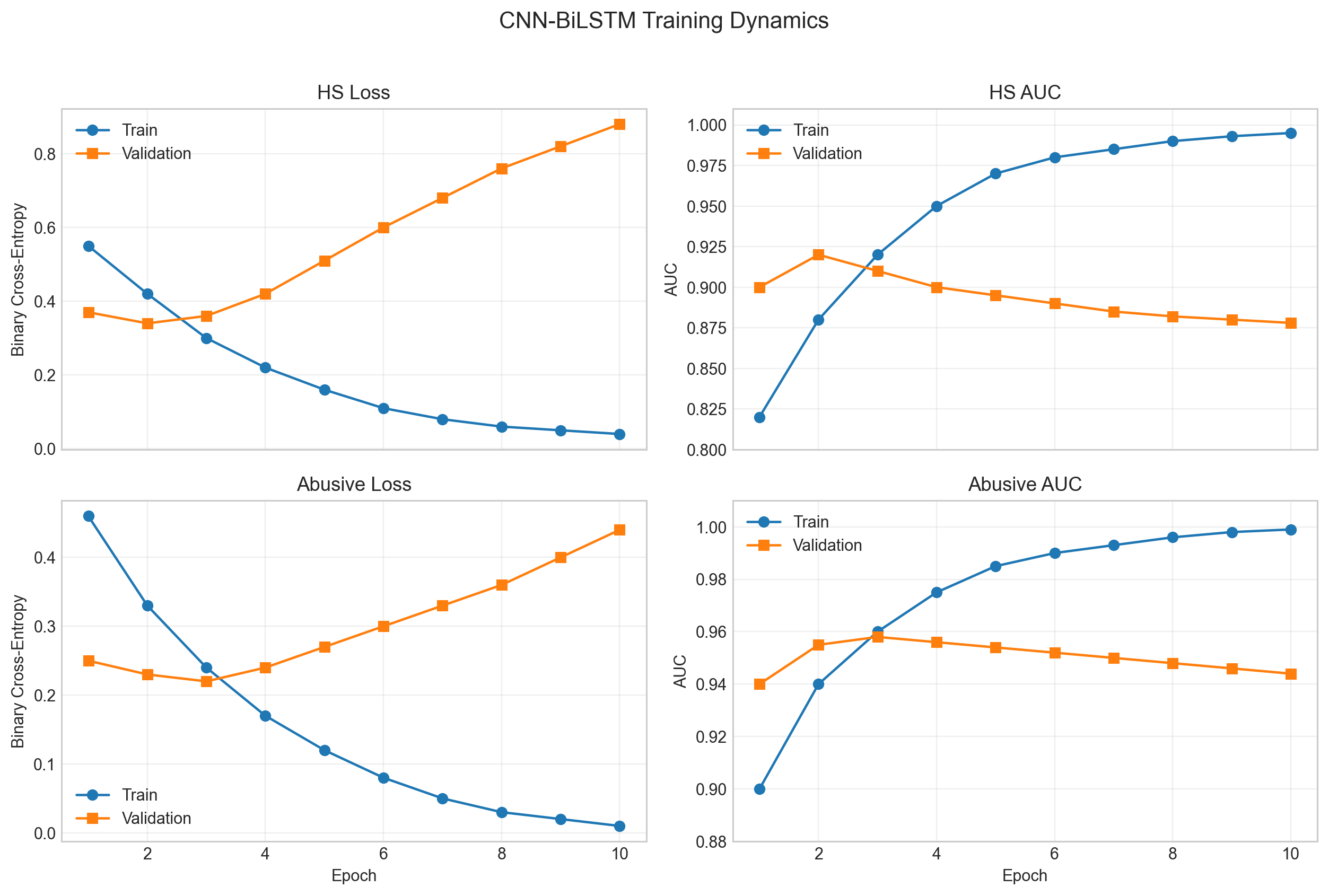}
    \caption{Training and validation loss and AUC curves for the HS and auxiliary abusive tasks.}\label{fig:training_curves}
\end{figure}

Figure~\ref{fig:training_curves} explains the benchmark result more carefully. On the HS task, training loss drops quickly, but validation loss rises after the early epochs and validation AUC declines slightly. This pattern is consistent with overfitting after the best early epochs. The auxiliary abusive-label run is more stable, which supports the interpretation that abusive-language recognition is easier than the stricter HS decision. The curves therefore suggest that future gains on HS are more likely to come from earlier stopping, stronger regularization, and better generalization control than from simply training longer.

The curve contrast is also useful for interpreting why the neural branch still beats the strongest PyCaret model. The CNN-BiLSTM clearly learns discriminative patterns quickly, which means the benchmark advantage is not merely a consequence of longer training or a larger parameter budget. At the same time, the widening train--validation gap indicates that the architecture is strong but not yet fully optimized for generalization. This makes the reported result conservative in an important sense: even with visible overfitting pressure, the neural branch still surpasses the best conventional comparator.

From an operational perspective, this matters because it separates representation quality from training stability. The present architecture already captures useful local and contextual cues, but the learning curves show that model selection should be tied to the best validation epoch rather than to the last epoch. In other words, the benchmark does not simply show that CNN-BiLSTM is better; it also shows how that advantage should be managed in practice through early stopping, stronger regularization, and threshold tuning.

\subsection{Error Analysis}
\begin{figure}[t]
    \centering
    \includegraphics[width=0.68\linewidth]{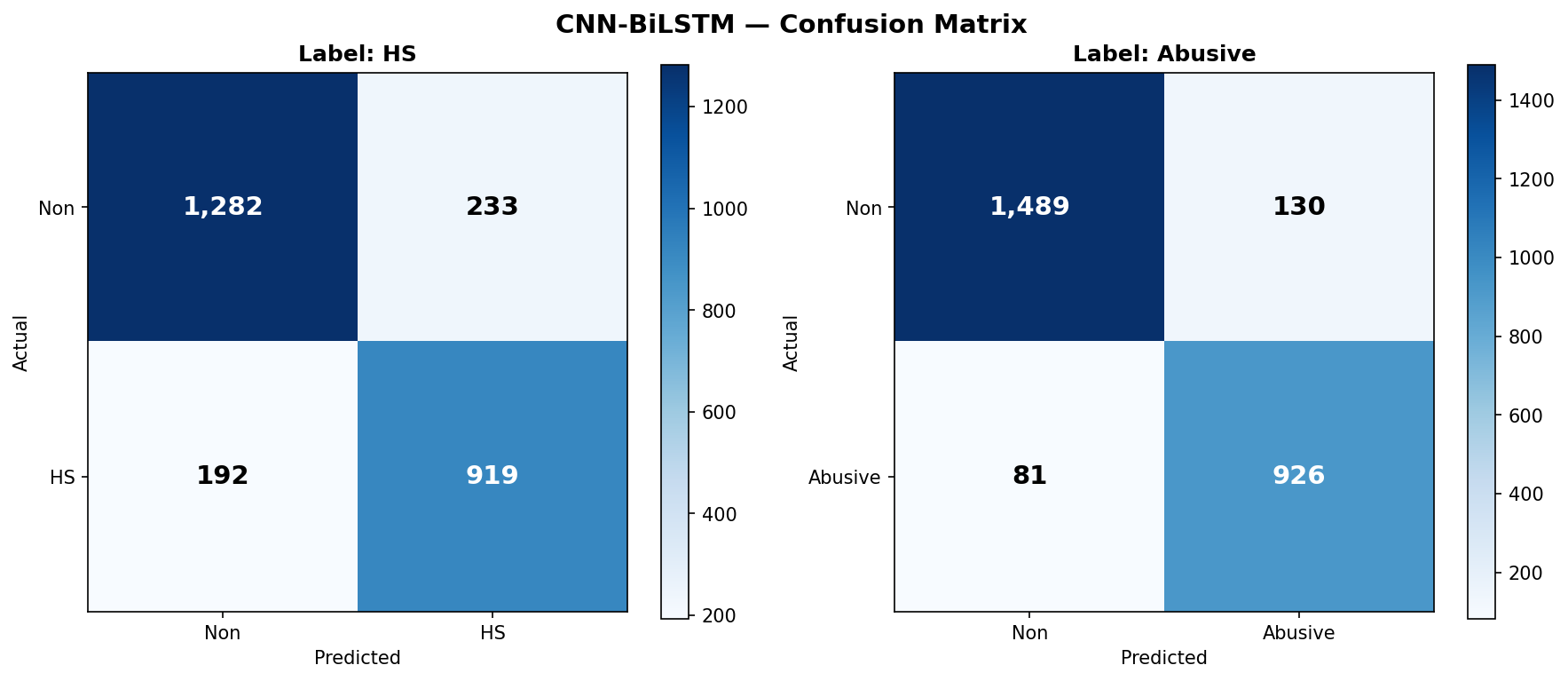}
    \caption{Confusion matrices for the HS and auxiliary abusive tasks.}\label{fig:confusion}
\end{figure}

Figure~\ref{fig:confusion} provides complementary error analysis. For the HS task, the model yields 1{,}282 true negatives, 919 true positives, 233 false positives, and 192 false negatives. The two error types are not extremely imbalanced, which means the classifier is somewhat aggressive but not unusably one-sided. The remaining mistakes are consistent with borderline tweets in which hostility is implied through group reference, short contextual composition, or weakly lexicalized intent rather than through an obvious abusive token.

The auxiliary abusive-label confusion matrix is more stable, with fewer total errors and a more symmetrical profile. This reinforces the view that explicit toxic lexicality is easier for the model family than the narrower HS decision boundary. In practical terms, the next improvements for HS are likely to come from better handling of context-sensitive borderline cases rather than from adding more lexical cues alone.

The error profile is also useful for thinking about moderation trade-offs. A system that raises recall only by sharply inflating false positives would be difficult to justify operationally, whereas a system that reduces false positives at the cost of missing too many hateful tweets would be equally problematic. The present HS confusion matrix suggests a more balanced regime: the classifier is somewhat aggressive, but the two error types remain close enough that threshold calibration and human review remain feasible strategies for deployment-oriented use.

\subsection{Study Scope and Practical Implications}
The findings should be read within the stated scope of the benchmark: one released dataset, one HS-focused task definition, one shared preprocessing pipeline, and one held-out split. For that reason, the reported numbers are best interpreted as evidence about this benchmark configuration rather than as universal claims about every form of Indonesian online hate speech.

Within that scope, the results still have practical value. PyCaret AutoML remains useful when researchers need rapid model screening, lower computational cost, and a transparent conventional benchmark. CNN-BiLSTM is the better choice when the primary objective is stronger predictive balance on the HS boundary and when the added modelling complexity is acceptable. The benchmark therefore supports a pragmatic reading of the title-level comparison: PyCaret is the stronger benchmarking framework for fast conventional screening, while CNN-BiLSTM is the stronger end model for the reported dataset and task.

This distinction is especially relevant in practice. Conventional models managed through PyCaret remain attractive for quick experimentation, reproducible baselining, and settings in which compute cost or deployment simplicity is a primary concern. The neural branch becomes more attractive when contextual nuance matters enough to justify additional modelling complexity and when validation-driven training control is available.

\section*{Statement on AI-Assisted Development}
Generative AI tools were used in a limited supporting role for drafting, implementation troubleshooting, and language refinement. All research design, result verification, interpretation, and final approval remain the responsibility of the authors.

\section{Conclusion}
This paper presented a controlled comparison between a PyCaret AutoML branch and a CNN-BiLSTM branch for binary hate-speech detection on Indonesian Twitter. The benchmark was derived from the released corpus of \citet{ibrohim2019}, evaluated under shared preprocessing assumptions, and assessed using accuracy, precision, recall, and F1-score.

The main conclusion is clear. PyCaret AutoML provides a strong and methodologically useful conventional benchmark, with Random Forest as its strongest reported model. Nevertheless, CNN-BiLSTM is the best-performing approach in this benchmark, achieving 83.8\% accuracy and 81.2\% F1-score and outperforming PyCaret-RF by 6.6 and 4.2 points, respectively. The most plausible reason is architectural fit: CNN-BiLSTM is better suited to short Indonesian Twitter text because it captures both local lexical patterns and short-range contextual interaction.

At the same time, the comparison should not be read as a dismissal of the conventional branch. The PyCaret workflow remains valuable precisely because it shows that the neural gain persists even after the classical baseline has been organized systematically and strengthened with lexical and abusive-cue features. Future work can strengthen this benchmark through richer AutoML search spaces, stronger regularization, threshold calibration, and transformer-based Indonesian models such as IndoBERT and IndoBERTweet~\citep{koto2020,koto2021}.

\bibliographystyle{IEEEtranN}
\bibliography{references}

@inproceedings{ibrohim2019,
  title={Multi-label hate speech and abusive language detection in {I}ndonesian {T}witter},
  author={Ibrohim, M. Okky and Budi, Indra},
  booktitle={Proceedings of the Third Workshop on Abusive Language Online},
  pages={46--57},
  year={2019},
  publisher={Association for Computational Linguistics},
  doi={10.18653/v1/W19-3506},
}

@article{pamungkas2024,
  title={Enhancing hate speech detection in {I}ndonesian using abusive words lexicon},
  author={Pamungkas, Endang Wahyu and Purworini, Dian and Putri, Divi Galih Prasetyo and Akhtar, Sohail},
  journal={Indonesian Journal of Electrical Engineering and Computer Science},
  volume={33},
  number={1},
  pages={450--462},
  year={2024},
  doi={10.11591/ijeecs.v33.i1.pp450-462},
}

@inproceedings{ramos2003,
  title={Using {TF-IDF} to determine word relevance in document queries},
  author={Ramos, Juan},
  booktitle={Proceedings of the First Instructional Conference on Machine Learning},
  pages={133--142},
  year={2003}
}

@inproceedings{kim2014,
  title={Convolutional neural networks for sentence classification},
  author={Kim, Yoon},
  booktitle={Proceedings of the 2014 Conference on Empirical Methods in Natural Language Processing (EMNLP)},
  pages={1746--1751},
  year={2014},
  publisher={Association for Computational Linguistics},
  doi={10.3115/v1/D14-1181},
}

@article{hochreiter1997,
  title={Long short-term memory},
  author={Hochreiter, Sepp and Schmidhuber, J{"u}rgen},
  journal={Neural Computation},
  volume={9},
  number={8},
  pages={1735--1780},
  year={1997},
  doi={10.1162/neco.1997.9.8.1735},
}

@inproceedings{koto2020,
  title={{I}ndo{LEM} and {I}ndo{BERT}: A benchmark dataset and pre-trained language model for {I}ndonesian {NLP}},
  author={Koto, Fajri and Rahimi, Afshin and Lau, Jey Han and Baldwin, Timothy},
  booktitle={Proceedings of the 28th International Conference on Computational Linguistics},
  pages={757--770},
  year={2020},
  publisher={International Committee on Computational Linguistics},
  doi={10.18653/v1/2020.coling-main.66},
}

@inproceedings{koto2021,
  title={{I}ndo{BERT}weet: A pretrained language model for {I}ndonesian {T}witter with effective domain-specific vocabulary initialization},
  author={Koto, Fajri and Lau, Jey Han and Baldwin, Timothy},
  booktitle={Proceedings of the 2021 Conference on Empirical Methods in Natural Language Processing},
  pages={10660--10668},
  year={2021},
  publisher={Association for Computational Linguistics},
  doi={10.18653/v1/2021.emnlp-main.833},
}

@book{hutter2019,
  title={Automated Machine Learning: Methods, Systems, Challenges},
  editor={Hutter, Frank and Kotthoff, Lars and Vanschoren, Joaquin},
  year={2019},
  publisher={Springer},
  doi={10.1007/978-3-030-05318-5},
}

@manual{pycaretdocs,
  title={PyCaret Documentation: Training Functions},
  author={{PyCaret}},
  year={2026},
  note={Official documentation, accessed May 2, 2026},
  url={https://pycaret.gitbook.io/docs/get-started/functions/train},
}

@article{schuster1997,
  title={Bidirectional recurrent neural networks},
  author={Schuster, Mike and Paliwal, Kuldip K.},
  journal={IEEE Transactions on Signal Processing},
  volume={45},
  number={11},
  pages={2673--2681},
  year={1997},
  doi={10.1109/78.650093},
}

\end{document}